\documentclass[lettersize,journal]{IEEEtran}
\usepackage{amsmath,amsfonts}
\usepackage{algorithmic}
\usepackage{algorithm}
\usepackage{array}
\usepackage[caption=false,font=normalsize,labelfont=sf,textfont=sf]{subfig}
\usepackage{textcomp}
\usepackage{stfloats}
\usepackage{url}
\usepackage{verbatim}
\usepackage{graphicx}
\usepackage{cite}
\usepackage{pifont}
\usepackage{booktabs}
\usepackage{multirow}
\usepackage[table,xcdraw]{xcolor}
\usepackage[colorlinks,
            linkcolor=blue,
            anchorcolor=blue,
            urlcolor=blue,
            citecolor=blue,
            ]{hyperref}
\usepackage[switch]{lineno}
\begin{document}
\title{Retinex-RAWMamba: Bridging Demosaicing and Denoising for Low-Light RAW Image Enhancement}

\author{Xianmin Chen, Longfei Han$^\dag$, Peiliang Huang$^\dag$, Xiaoxu Feng, Dingwen Zhang, Junwei Han,~\IEEEmembership{Fellow,~IEEE,}
\thanks{This work was supported by the National Natural Science Foundation of China (No. U24A20341, No. 62202015), Anhui Provincial Key R\&D Programmes (2023s07020001) and Anhui Province Postdoctoral Researchers Research Grant Program (RS25BH004). (Corresponding author: Longfei Han and Peiliang Huang.)\\
X. Chen is with Institute of Advanced Technology, University of Science and Technology of China, Hefei, 230026, China
(E-mail: yicarlos@mail.ustc.edu.cn)\\
L. Han is with School of Computer and Artificial Intelligence, Beijing Technology and Business University, Beijing, 102488, China (E-mail: draflyhan@gmail.com)\\
P. Huang and X.Feng are with Institute of Artificial Intelligence, Hefei Comprehensive National Science Center, Hefei, 230088, China and University of Science and Technology of China, Hefei, 230026, China(E-mail: peilianghuang2017@gmail.com, fengxiaox@mail.nwpu.edu.cn)\\
D. Zhang and J. Han are with School of Automation, Northwestern Polytechnical University, Xian, 710129, China (E-mail: zhangdingwen2006yyy@gmail.com, junweihan2010@gmail.com))
}
\thanks{Manuscript received April 19, 2021; revised August 16, 2021.}}

\markboth{Journal of \LaTeX\ Class Files,~Vol.~14, No.~8, August~2021}%
{Shell \MakeLowercase{\textit{et al.}}: A Sample Article Using IEEEtran.cls for IEEE Journals}

\IEEEpubid{\begin{minipage}{\textwidth}\ \\[12pt] \centering
		Copyright \copyright 20xx IEEE. Personal use of this material is permitted. However, permission to use this material for any other purposes must \\be obtained from the IEEE by sending an email to pubs-permissions@ieee.org.
\end{minipage}}

\maketitle
\begin{abstract}
Low-light image enhancement, particularly in cross-domain tasks such as mapping from the raw domain to the sRGB domain, remains a significant challenge. Many deep learning-based methods have been developed to address this issue and have shown promising results in recent years. However, single-stage methods, which attempt to unify the complex mapping across both domains, leading to limited denoising performance. In contrast, existing two-stage approaches typically overlook the characteristic of demosaicing within the Image Signal Processing (ISP) pipeline, leading to color distortions under varying lighting conditions, especially in low-light scenarios. To address these issues, we propose a novel Mamba-based method customized for low light RAW images, called RAWMamba, to effectively handle raw images with different CFAs. Furthermore, we introduce a Retinex Decomposition Module (RDM) grounded in Retinex prior, which decouples illumination from reflectance to facilitate more effective denoising and automatic non-linear exposure correction, reducing the effect of manual linear illumination enhancement. By bridging demosaicing and denoising, better enhancement for low light RAW images is achieved. Experimental evaluations conducted on public datasets SID and MCR demonstrate that our proposed RAWMamba achieves state-of-the-art performance on cross-domain mapping. The code is available at \url{https://github.com/Cynicarlos/RetinexRawMamba}.
\end{abstract}


\section{Introduction}
\IEEEPARstart{E}xisting deep learning methods, particularly those focused on low-light enhancement tasks, primarily operate in the sRGB domain. However, RAW images typically possess a higher bit depth than their RGB counterparts, meaning they retain a greater amount of original detail. Consequently, processing from RAW to RGB is often more effective. However, RAW and RGB are distinct domains with image processing algorithms tailored to their specific characteristics. For instance, in the RAW domain, algorithms prioritize denoising, whereas in the RGB domain, they focus on color correction. This difference often renders single-stage end-to-end methods \cite{DID, SGN, RRT} ineffective.
\begin{figure}
    \centering
    \includegraphics[width=\columnwidth]{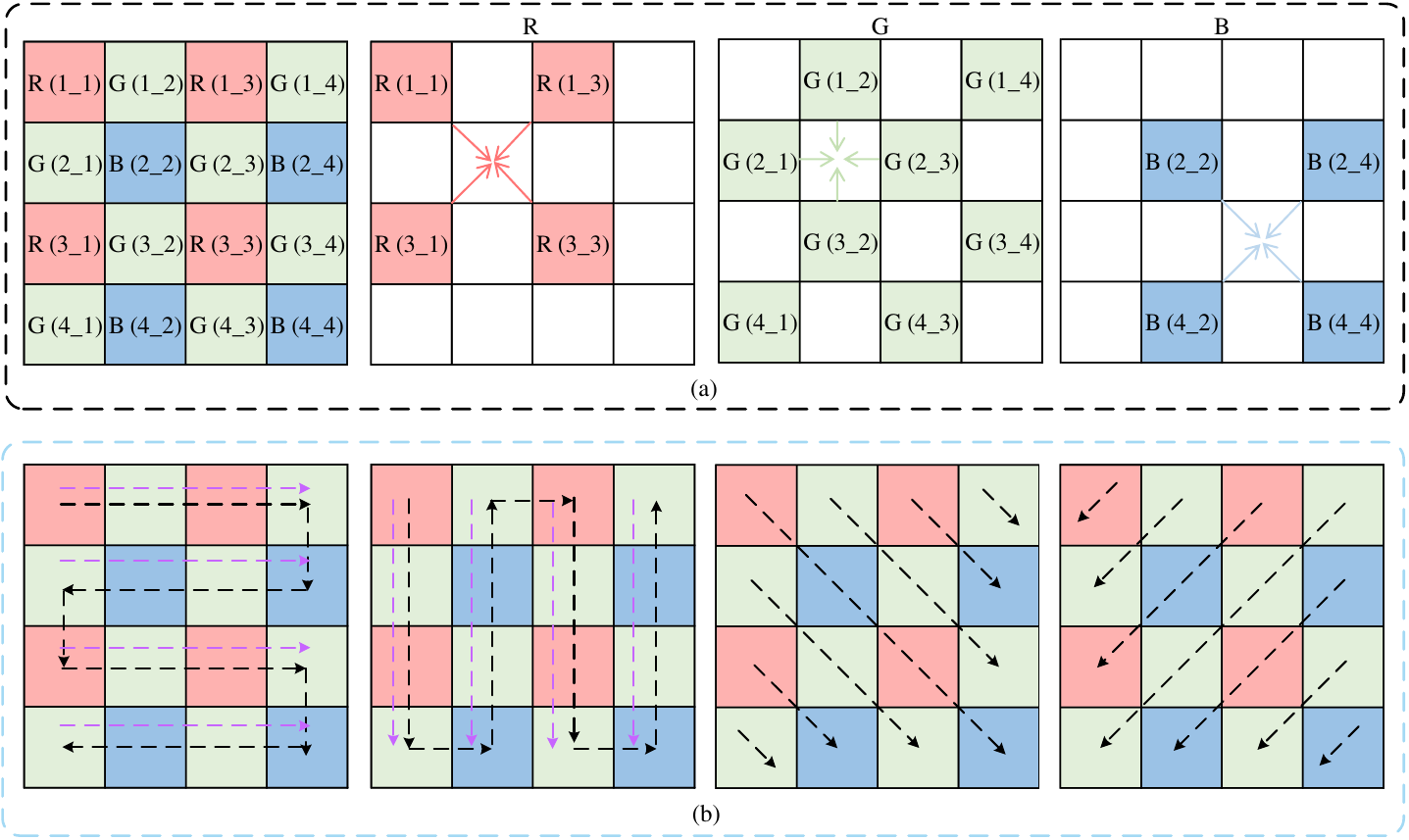}
    \caption{(a) A kind of demosaicing interpolation for RGGB Bayer Pattern and (b) the scanning in RAWMamba (black dashed line) and naive Mamba (purple dashed line). Note that only four directions of RAWMamba are drawn, reversing them gives four more directions, eight in all.}
    \label{fig:demosaicing}
\end{figure}

Demosaicing algorithms play a crucial role in converting RAW image to sRGB, with most traditional methods relying on proximity interpolation. Although some researchers have explored CNN-based approaches \cite{LDC, MCR} to map noisy RAW images to clean sRGB outputs, the limited receptive field inherent in convolutional networks often hampers their effectiveness in demosaicing tasks. To address this, Vision Transformers (ViTs) have been employed to expand the receptive field, but the attention mechanisms in ViTs are computationally intensive. The introduction of Mamba provides a more efficient balance between these trade-offs \cite{vmamba, vmambaIR, mambaIR}. However, existing Mamba scanning mechanisms do not adequately address the diverse characteristics of RAW images with different Color Filter Arrays (CFAs), highlighting the need for Mamba scanning methods specifically tailored to various CFAs. 

\IEEEpubidadjcol

Hence, we design a novel Mamba scanning mechanism for RAW format image(RAWMamba), which has a global receptive field and an attention mechanism with linear complexity that can better adapt to the data in this task. More importantly, as shown in Fig. \ref{fig:demosaicing} (b), naive Mamba scanning mechanism do not consider imaging properties, leading to limitations in feature extraction with CFA. In contrast, our RAWMamba introduces eight distinct scanning directions, fully accounting for all pixels in the immediate neighborhood of a given pixel while preserving the spatial continuity of the image. Specifically, the scanning directions encompass horizontal, vertical, oblique scanning from top left to bottom right, and oblique scanning from top right to bottom left. These four primary directions are mirrored to produce an additional four directions, resulting in a total of eight scanning directions.

Additionally, previous methods \cite{SID, DNF} for processing short-exposure RAW images often rely on a simple linear multiplication of a prior for exposure correction. Specifically, short-exposure RAW images, which contain significant noise, are multiplied by the exposure time ratio of the corresponding long-exposure image. This approach assumes uniform exposure across the image, which is often unrealistic and can result in sub-optimal denoising and inaccurate brightness. By leveraging the success of the Retinex theory in low-light enhancement tasks for RGB images \cite{retinex1,retinex2,retinex3}, we introduce a Retinex-based dual-domain auxiliary exposure correction method, namely Retinex Decomposition Module (RDM), which decouples illumination and reflection and realize automatic nonlinear exposure correction. At the same time, we efficiently fuse the generated priors based on multi-scale fusion strategy\cite{2-6,2-7,2-8,2-9,2-10,2-11} to achieve more efficient denoising effect and more accurate brightness correction. Furthermore, given the significant differences in noise distribution between different RAW domain and sRGB domain, we build upon the idea of decoupling the task into two sub-tasks: denoising on the RAW domain and cross-domain mapping.

\begin{figure*}
    \centering
    \includegraphics[width=\linewidth]{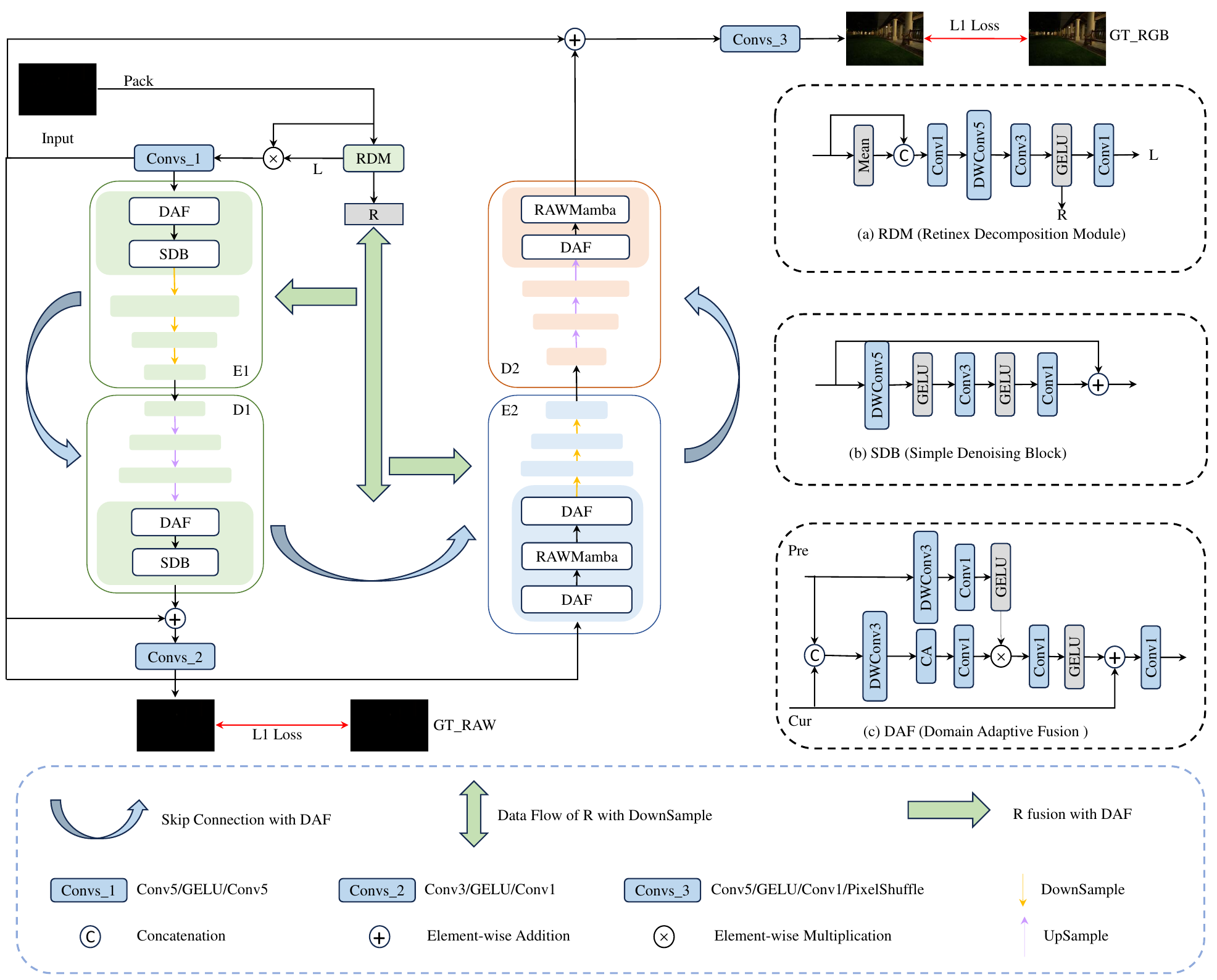}
    \caption{The overall architecture of our proposed Retinex-RAWMamba and (a) Retinex Decomposition Module, (b) Simple Denoising Block and (c) Domain Adaptive Fusion}
    \label{fig:overall}
\end{figure*}

In general, we propose a Retinex-based decoupling network (Retinex-RAWMamba) for RAW domain denoising and low-light enhancement shown in Fig. \ref{fig:overall}. Our method decouples the tasks of denoising and demosaicing into two distinct sub-tasks, effectively mapping noisy RAW images to clean sRGB images. Specifically, for demosaicing sub-task, we introduce RAWMamba to fully consider all pixels in the immediate neighborhood of a certain pixel by utilizing eight direction mechanism. For the denoising sub-task, we propose the Retinex Decomposition Module, which enhances both denoising performance and brightness correction. Additionally, we introduce a dual-domain encoding stage enhance branch designed to leverage the meticulously preserved detail features from the raw domain, thereby compensating for the information loss that occurs during the denoising phase. Finally, We observe that the Gated Fusion Module (GFM) used in prior works led to unstable training within our framework. To address this, Domain Adaptive Fuse (DAF)is proposed to perform adaptive feature fusion with better stability and efficiency.

Our main contributions are summarized as follows:
\begin{itemize}
    \item We propose a Retinex-based decoupling Mamba network for RAW domain denoising and low-light enhancement (Retinex-RAWMamba). To our best knowledge, this is first attempt to introduce Mamba mechanism into low-light RAW image task.
    \item We design a novel eight-direction Mamba scanning mechanism, to thoroughly account for the intrinsic properties of RAW images, and develop a Retinex Decomposition Module to bridging denoising capabilities and exposure correction.
    \item We evaluate the proposed method on two benchmark datasets quantitatively and qualitatively. The comprehensive experiments show that the proposed method outperforms other state-of-the-art methods in PSNR, SSIM and LPIPS with a comparable number of parameters.
\end{itemize}

\section{Related Work}
\subsection{Low Light Enhancement on Raw Domain}
Raw image consists much more details than its corresponding RGB image, and it would be better to enhance the low light image on Raw domain than RGB domain, like some homomorphic filtering methods \cite{1-1, 1-2, 1-3}, which are a great inspiration for us. For the Raw domain low-light enhancement task, researchers have proposed some innovative approaches. Since the task can be split into two sub-tasks, RAW denoising and cross domain mapping. For example, on the raw domain denoising task, there are noise modeling with deep learning methods \cite{raw4,raw5,raw6,raw7,raw8,raw9}, which ultimately compute evaluation metrics on the raw domain. After the release of the SID public dataset by Chen at al.\cite{SID} 2018, researchers have proposed many works that address both tasks simultaneously. These works can be further categorized into single-stage approaches and multi-stage approaches. Single-stage methods \cite{SID,zamir2019learning} aims to map noisy raw to clean sRGB by training a single model. For instance, SID \cite{SID} only used a simple UNet to accomplish this task. DID \cite{DID} proposed a deep neural network based on residual learning for end-to-end extreme low-light image denoising. SGN \cite{SGN} introduced a self-guided network, which adopted a top-down self-guidance architecture to better exploit image multi-scale information. 

Since the ISP undergoes many nonlinear transformations, it is still difficult to learn for a single neural network, and it can only be realized by piling up a large number of parameters, which leads to inefficiencies, and thus multi-stage methods came into being. Multi-stage methods \cite{RRENet, LDC, EEMEFN} achieve better results by decoupling the tasks, this idea effectively reduces the ambiguity between different domains. For instance, Huang et al. \cite{RRENet} proposed intermediate supervision on the raw domain, while Dong et al. \cite{MCR} did that on the monochrome domain. DNF \cite{DNF} introduced a decoupled two-stage net with weight-shared encoder to reduce the number of parameters while achieving good results. However, the weight-sharing module used across both domains may introduce cross-domain ambiguity, resulting in suboptimal performance.

\subsection{Deep Learning for ISP}
The motivation for replacing hardware-based ISP systems with deep learning solutions stems from their superior capability in reconstructing lost image information while mitigating cumulative processing errors inherent in traditional multi-stage ISP pipelines \cite{isp0}. Recent advancements in this field have demonstrated significant progress through diverse methodological innovations \cite{isp9, isp10}. CycleISP framework proposed by Zamir et al. \cite{isp2} features a bidirectional architecture containing complementary RGB2RAW and RAW2RGB conversion branches, enhanced by an adaptive color correction module to simulate camera imaging pipelines and generate paired data for dual-domain denoising. For demosaicing optimization, Xu et al. \cite{isp5} developed a hierarchical processing architecture called DemosaicFormer, implementing coarse reconstruction followed by pixel-level refinement. Addressing cross-device adaptability, Perevozchikov et al. \cite{isp6} pioneered an unpaired learning paradigm for RAW-to-RAW translation across heterogeneous camera sensors, enabling flexible deployment of neural ISPs on unseen devices. MetaISP framework by Souza et al. \cite{isp7} introduced metadata-aware domain adaptation, leveraging EXIF parameters and illuminant estimation to achieve cross-device characteristic translation. In computational efficiency optimization, Guan et al. \cite{isp8} innovated a grouped deformable convolution mechanism for joint denoising and demosaicing, strategically allocating independent offset parameters across kernel groups to balance accuracy and latency. Collectively, these advancements validate the viability of deep learning-based ISP solutions through comprehensive technical explorations spanning data synthesis, architectural design, and deployment optimization.

\subsection{Mamba in Vision Task}
State Space Models (SSM) are recently introduced to deep learning since they can effectively model long range dependencies. For instance, \cite{gu2022efficiently} proposes a Structured State-Space Sequence (S4) model and recently, \cite{gu2023mamba} proposes Mamba, which outperforms Transformers at various sizes on large-scale real data and enjoys linear scaling in sequence length. In addition to Mamba's great work on NLP tasks, researchers have also made many attempts and achieved good results on visual tasks\cite{2-1,2-2}, such as classification \cite{cla1,cla2}, segmentation \cite{seg1,seg2,seg3,seg4,seg5,2-5}, anomaly detection\cite{2-3}, point cloud learning\cite{2-4}, generation \cite{gen1,gen2}, and image restoration \cite{mambaIR,vmambaIR,ushaped_vision_mamba,efficientvmamba,retinexmamba}. \\
EfficientVMamba \cite{efficientvmamba} presents the Efficient 2D Scanning (ES2D) method, utilizing atrous sampling of patches on the feature map to speed up training. VMamba \cite{vmamba} incorporates a Cross-Scan Module (CSM), which converts the input image into sequences of patches along the horizontal and vertical axes, and it enables the scanning of sequences in four distinct directions. That is, each pixel integrates information from the four surrounding pixels. VMambaIR \cite{vmambaIR} proposes omni selective scan mechanism to overcome the unidirectional modeling limitation of SSMs by efficiently modeling image information flows in all six directions for RGB image images, which added two more directions along the channel dimension. In contrast, we combine the characteristics of AI ISP task, proposing the eight direction scanning mechanism and Retinex decomposition module to overcome the uneven lighting of the low light RAW images. FreqMamba \cite{freqmamba} introduces complementary triple interaction structures including spatial Mamba, frequency band Mamba, and Fourier global modeling, which utilizes the complementarity between Mamba and frequency analysis for image deraining.
Similarly, WalMaFa\cite{3-3} proposes a novel Wavelet-based Mamba with Fourier adjustment model. RetinexMamba\cite{retinexmamba} directly integrates Mamba into RetinexFormer\cite{retinexformer} for low light RGB image enhancement without any changes about the Mamba itself. Li et al.\cite{3-2} combines contrastive learning and Mamba to achieve semi-supervised learning. However, most of the existing scanning mechanism in the mentioned Mamba have limitations. One direction scanning in the original Mamba \cite{gu2023mamba} that is used for sequence prediction task does usually not do well in vision tasks, since the image has two dimensions and each pixel are usually related to its surrounding pixels instead of only its next one as the sequence. Therefore, most vision Mamba adopt the four directions scanning mechanism. Li et al. \cite{3-4} utilizes four directions scanning strategy for underwater image enhancement. Nevertheless, it just consider the up, down, left and right pixels and ignore the spatial continuity, which is not the best approach in our task. Therefore, RetinexRawMamba is proposed to address this issue by considering eight scanning directions. And to address the uneven light of the low light RAW image, we introduced the retinex decomposition module to estimate the light components to adaptively correct the light and color by multi-scale fusion, which can also complement mamba's ability to capture local information.

\section{Method}
\subsection{Preliminaries}
\subsubsection{State Space Model (SSM)}
SSM is a linear time-invariant system that maps input $x(t)\in {\mathbb{R}}^L$ to output $y(t) \in {\mathbb{R}}^L$. SSM can be formally represented by a linear ordinary differential equation (ODE),
\begin{equation}  
    \begin{aligned}  
        h'(t) &= \mathbf{A}h(t) + \mathbf{B}x(t), \\  
        y(t) &= \mathbf{C}h(t) + \mathbf{D}x(t)  
    \end{aligned}
\label{eq1}
\end{equation} 
SSM is continuous-time model, presenting significant challenges when integrated into deep learning algorithms. To address this issue, discretization becomes a crucial step. Denote $\Delta$ as the timescale parameter. The zero-order hold (ZOH) rule is usually used for discretization to convert continuous parameters $\mathbf{A}$ and $\mathbf{B}$ in Eq. \ref{eq1} into discrete parameters $\overline{\mathbf{A}}$ and $\overline{\mathbf{B}}$. Its definition is as follows:
\begin{equation}
    \begin{aligned}
        \overline{\mathbf{A}} &= exp(\Delta \mathbf{A}), \\
        \overline{\mathbf{B}} &= (\Delta \mathbf{A})^{-1}(exp(\Delta\mathbf{A})-\mathbf{I})\cdot\Delta\mathbf{B}
    \end{aligned}
\label{eq2}
\end{equation}
After the discretization of A, B, the discretized version of Eq. \ref{eq1} using a step size $\Delta$ can be rewritten as:
\begin{equation}
    \begin{aligned}
        h_k &= \overline{\mathbf{A}}h_{k-1} + \overline{\mathbf{B}}x_k,\\
        y_k &= \mathbf{C}h_k + \mathbf{D}x_k
    \end{aligned}
\label{eq3}
\end{equation}
Finally, the models compute output through a global convolution as follows:
\begin{equation}
    \begin{aligned}
        \overline{\mathbf{K}} &= (\mathbf{C}\overline{\textbf{B}},\mathbf{C}\overline{\mathbf{AB}},...,\mathbf{C}{\overline{\mathbf{A}}}^{L-1}\overline{\mathbf{B}}) \\
        \mathbf{y} &= \mathbf{x} \ast \overline{\mathbf{K}}
    \end{aligned}
\end{equation}
where L is the length of the input sequence \textbf{x}, and $\overline{\mathbf{K}} \in {\mathbb{R}}^L$ is a structured convolutional kernel.
\subsection{Overall Pipeline}
The overall pipeline is shown in Fig. \ref{fig:overall}. First, we preprocess the low-exposure noisy single-channel raw image by multiplying it with the exposure time ratio of the long-exposure ground truth (GT). Then, based on the Color Filter Array (CFA) pattern, we pack it into a multi-channels input. Specifically, for Bayer format, we pack the input $\textbf{X} \in \mathbb{R}^{H \times W \times 1}$ into four channels input $\textbf{X} _{packed} \in \mathbb{R}^{\frac{H}{2} \times \frac{W}{2} \times 4}$; for XTrans format, we pack the input into nine channels input $\textbf{X} _{packed} \in \mathbb{R}^{\frac{H}{3} \times \frac{W}{3} \times 9}$. Both stages of Retinex-RAWMamba are built upon the UNet-based \cite{unet} encoder-decoder architecture.
While the U-Net architecture remains a parsimonious choice for dense prediction tasks, our research introduces a domain-specific architectural innovation that transcends conventional designs: the first-ever integration of Mamba with Retinex-guided multi-scale prior fusion for low-light RAW image enhancement. This approach addresses a critical research gap—existing Mamba applications in 2024 have overlooked RAW data's unique challenges, as validated by our comprehensive literature survey (Section \uppercase\expandafter{\romannumeral2}.C). Furthermore, The experimental results (Section \uppercase\expandafter{\romannumeral4}.B) also prove that this architecture design has a significant performance improvement over the baseline architecture.\\
The first stage of the overall framework is dedicated to raw domain denoising. Initially, The Retinex Decomposition Module (RDM) processes the input to generate two feature maps $\textbf{L}$ and $\textbf{R}$, $\textbf{L}$ will be multiplied by the original input to obtain $\textbf{X}_{in}$ and $\textbf{R}$ will be used later. 
$\textbf{X}_{in}$ will pass the denoising stage and generate the first output $\textbf{O}_{1} \in \mathbb{R}^{H \times W \times C_{in}}$. Then $\textbf{X}_{in}$ will pass the demosaicing stage to generate the second output $\textbf{O}_{2} \in \mathbb{R}^{H \times W \times 3}$. 
The overall loss function is then calculated against both ground truth RAW and ground truth RGB images, providing the supervision signal for both domains and guiding the optimization of whole model.
\begin{figure*}
    \centering
    \includegraphics[width=\linewidth]{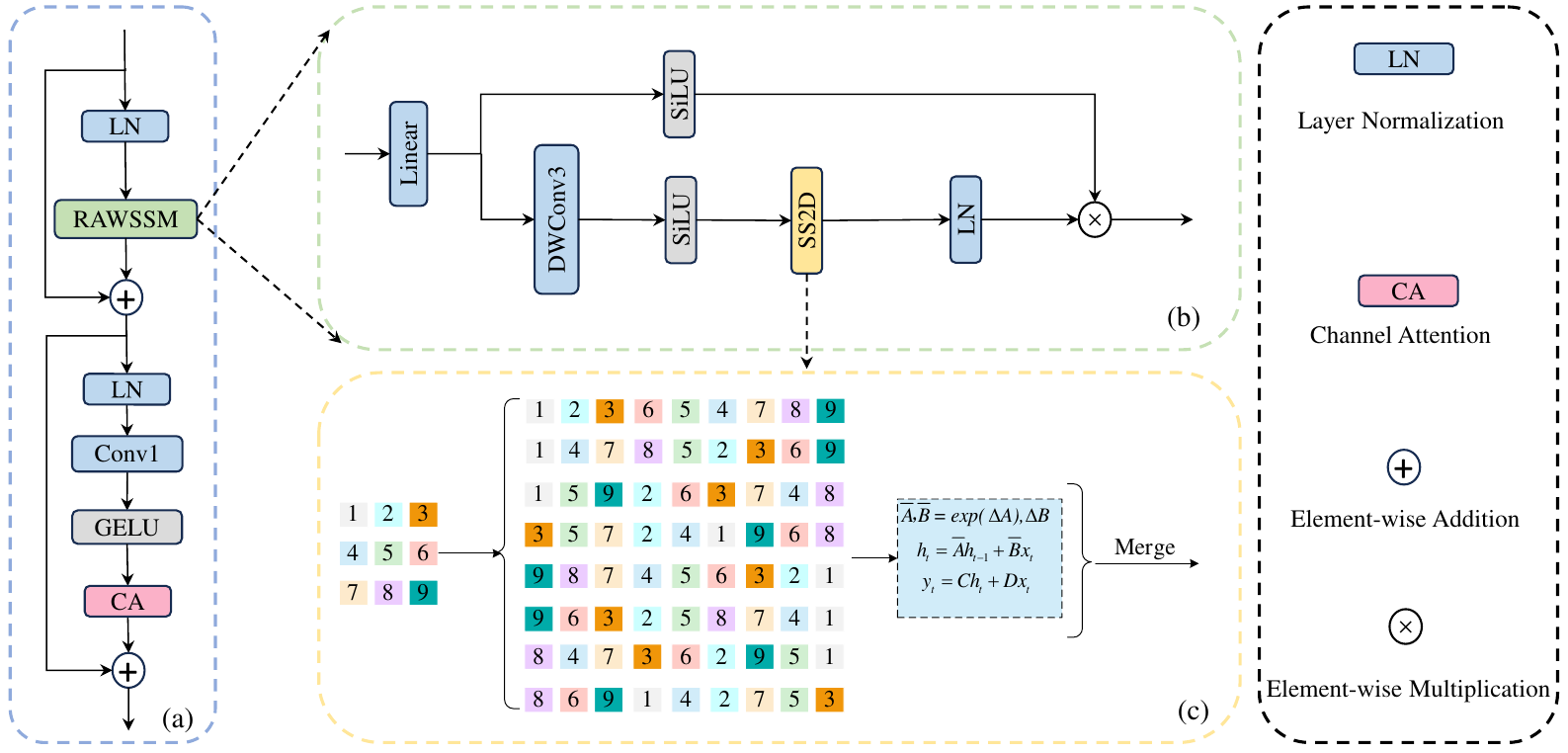}
    \caption{Details of (a) RAWMamba, (b) RAWSSM and (c) SS2D}
    \label{fig:mamba}
\end{figure*}
\subsection{RAWMamba}
The details of RAWMamba and is shown in Fig. \ref{fig:mamba} (a). The RAWSSM leverages the naive visual mamba in MambaIR \cite{mambaIR}, with an innovative scanning mechanism. In the ISP process from Raw to RGB, proximity interpolation is commonly employed for demosaicing and often involves considering all eight closely connected locations around a given position, and Fig. \ref{fig:demosaicing} (a) gives an example with Bayer pattern raw image, (b) shows the scanning in RAWMamba (black
dashed line) and naive Mamba (purple dashed line). The naive scanning method fails to consider the continuity of scanning, resulting in a lack of continuity between the end of each row/column and its bottom/right side. This leads to gaps in image semantics, which hinders image reconstruction. To address this issue, we propose using a Z-scan. That is when the scan reaches the end of each row/column, the reverse scan starts from the next row/column immediately adjacent to the last pixel. However, there are still limitations with this scanning method as it does not take into account all eight surrounding pixels when certain pixels are close to each other at the top, bottom, left, and right positions. Taking into consideration the characteristics of this task, we introduce Eight direction Mamba. 

The detail of the our proposed scan mechanism is shown in Fig. \ref{fig:mamba} (c). 
We will firstly obtain eight directions scanning features, which is $\{\textbf{\textit{F}}_i \in {\mathbb{R}}^{C\times HW}, i = 1,2,...8\}$. At this point, the scanning of the eight directions is completed. And after the SSM, we get $\{\overline{\textbf{\textit{F}}}_i \in {\mathbb{R}}^{C\times HW}, i = 1,2,...8\}$, we then merge them by adding them up and reshape these eight features to the original shape to get a single feature, that is,
\begin{equation}
    SS2D(\textbf{\textit{F}}) = Reshape(\sum_{i=1}^{8}\overline{\textbf{\textit{F}}}_i, (C,H,W))
\end{equation}

For the RAWSSM, given an input \textbf{\textit{X}}, it can be formulated as follows:
\begin{equation}
    \begin{aligned}
         x, z &= chunk(Linear(\textbf{\textit{X}}))\\
         x &= LN(SS2D(SiLU(Conv_3(x))))\\
         out &= x * SiLU(z) 
    \end{aligned}
\end{equation}
where out is the output of RAWSSM, LN is layer normalization, $Conv_3$ is the convolution operation with a kernel size of $3\times3$, SiLU is the activate function.

And for the proposed RAWMamba, given an input \textbf{\textit{X}}, it can be simply formulated as:
\begin{equation}
    \begin{aligned}
        t &= \alpha\textbf{\textit{X}} + RAWSSM(LN(\textbf{\textit{X}})))\\
        out &= \beta t + CA(GELU(Conv(LN(t))))
    \end{aligned}
\end{equation}
where, out is the output of RAWMamba, $\alpha$ and $\beta$ are parameters that can be learned, CA is channel attention.
\subsection{Retinex Decomposition and Dual-domain Encoding Stage Enhance Branch}
Low-light enhancement methods based on retinex theory have been successful in RGB domain \cite{retinex4,retinex5,retinexformer}, so we propose dual-domain Retinex Decomposition Module. Our RDM is inspired by RetinexFormer\cite{retinexformer}, which just use several convolutions to estimate the illumination map and the reflect map. The previous retinex-based methods are one stage network and the prior features are used in both encoding and decoding stage, but we remove two decoding stages prior fusion in our two stages network to reduce the computational complexity.
RDM can decompose image $\textbf{\textit{X}} \in {\mathbb{R}}^{H \times W \times C_{in}}$ into the reflection component $\textbf{\textit{R}} \in {\mathbb{R}}^{H \times W \times C}$ and the illumination component $\textbf{\textit{L}} \in {\mathbb{R}}^{H \times W \times C_{in}}$. The details of RDM are shown in Fig. \ref{fig:overall} (a). The module first takes the average value of the input image $\textbf{\textit{X}} \in {\mathbb{R}}^{H \times W \times C_{in}}$ in the channel dimension to obtain $\textbf{\textit{M}} \in {\mathbb{R}}^{H \times W \times 1}$, concatenates it in the channel dimension, and then passes several convolutions and a GELU activate function to obtain the first output $\textbf{\textit{R}} \in {\mathbb{R}}^{H \times W \times C}$, and then passes a $1\times1$ convolution to obtain the light map $\textbf{\textit{L}}$ , which will be multiplied by the original input to pre-adjust the light. Specifically,
\begin{equation}
    \begin{aligned}
        \textbf{\textit{L}} &= Convs_{1,5,3} \{cat[\textbf{\textit{X}}, mean(\textbf{\textit{X}}, dim=-1)]\}\\
        \textbf{\textit{R}} &= Conv_{1}(\textbf{\textit{L}})
    \end{aligned}
\end{equation} 
where cat refers to the concatenation of two feature maps on the channel dimension, $Convs_{1,5,3}$ donates a series of convolution with kernel size 1, 5 and 3.\\
\begin{figure*}[htb]
    \centering
    \includegraphics[width=\linewidth]{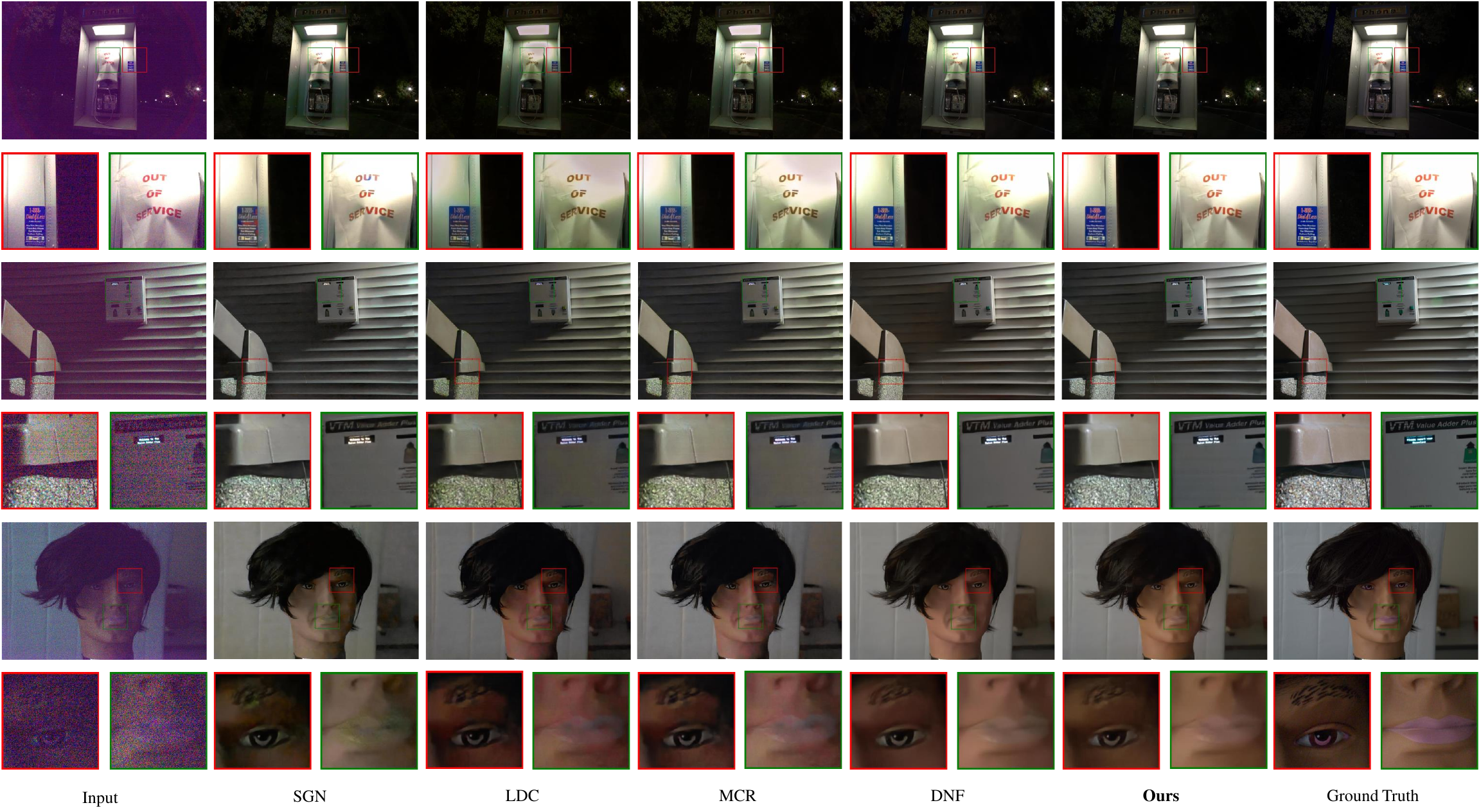}
    \caption{The visualization results between our method and the state-of-the-art methods, and the red and green box areas are cropped out for easy comparison (Zoom-in for best view).}
    \label{fig:visualization}
\end{figure*}
Considering that the feature $\textbf{\textit{R}}$ obtained from the RDM contains most of the details that could be lost after the first stage, we make full use of these features in both domains and in order to reduce the amount of calculation, we propose dual-domain encoding stage enhance branch, which does not be used in the decoding stages. Specifically, after obtaining \textbf{\textit{R}}, we will simply downsample it to get four feature maps at each layer, which are donated as \{$\textbf{\textit{R}}_{i}, i=1,2,3,4\}$, $\textbf{\textit{R}}_{i}$ is the $i_{th}$ layer light feature that will be fused later with DAF for auxiliary automatic exposure correction at layer $i$. At the $i_{th}$ layer of the encoding stage during the denoising phase, the denoising feature $\textbf{\textit{dn}}_{i}$ will firstly fused with $\textbf{\textit{R}}_{i}$ before passing the SDB. Similarly, at the $i_{th}$ layer of the encoding stage during the demosaicing phase. the the demosaicing feature $\textbf{\textit{dm}}_{i}$ will firstly fused with $\textbf{\textit{R}}_{i}$ before passing the RAWMamba. This method of performing fusion only in the encoding stage fully utilizes the features that are not lost to improve the ability to restore details while reducing the amount of calculation.

\subsection{Domain Adaptive Fusion} 
The details of DAF are shown in Fig. \ref{fig:overall} (c), previous feature map will be firstly concatenated with current feature map at the same level, and this result will be multiplied with previous feature map after the convolution, then it will pass through a convolution with a residual addition. And we can get the fused feature map after a final convolution. Specifically, for the two feature maps $\textbf{\textit{pre}} \in {\mathbb{R}}^{H \times W \times C}$ and $\textbf{\textit{cur}} \in {\mathbb{R}}^{H \times W \times C}$, they will be fused as follows:\\
\begin{equation}
    \begin{aligned}
        & \textbf{\textit{T}} = Conv_{3}(cat(\textbf{\textit{pre}},\textbf{\textit{cur}}))\\
        & \textbf{\textit{T}} = Conv_{1}(CA(\textbf{\textit{T}}))\\
        & \textbf{\textit{T}} = \textbf{T} \odot Conv_{1}(GELU(\textbf{\textit{pre}}))\\
        & \textbf{\textit{T}} = Conv_{1}(GELU(\textbf{\textit{T}}))\\
        & Out(\textbf{\textit{cur}}, \textbf{\textit{pre}}) = Conv_{1}(\textbf{\textit{T}}+\textbf{\textit{cur}})
    \end{aligned}
\end{equation}
\subsection{Loss Function}
Traditional low-level vision tasks generally use L1 Loss, and we also follow that, but our task involves different sub-tasks on two domains, Raw domain and sRGB domain, so we use L1 loss for each domain to better guide model learning. And the total loss can be expressed as follows: 
\begin{align}
    L_{total} &= \alpha L_{raw} + \beta L_{srgb} \notag\\
    &= \alpha {||\hat{Y}_{raw} - GT_{raw}||}_1 + \beta {||\hat{Y}_{srgb} - GT_{srgb}||}_1 \notag
\end{align}
where $Y_{raw}$ is the raw image after denoised, $Y_{srgb}$ is the sRGB image after the second stage, $GT_{srgb}$ is the sRGB image obtained from raw ground truth after post-processing by Rawpy as previous work did. And $\alpha$ and $\beta$ defaults to 1.0 in our experiments.
\begin{table*}
\centering
\renewcommand\arraystretch{1.5}
\caption{Quantitative results of RAW-based LLIE methods on the Sony and Fuji subsets of SID. The top-performing result is highlighted in \textbf{bold}, while the second-best is shown in \underline{underline}. Metrics marked with $\uparrow$ indicate that a higher value is better, and those marked with $\downarrow$ indicate that a lower value is better. `-' indicates the result is not available.}
\setlength\tabcolsep{6.8pt}{
\begin{tabular}{ccccccccccc}
\toprule[1.2pt]
\multirow{2}{*}{Category}     & \multirow{2}{*}{Method} & \multirow{2}{*}{Venue} & \multirow{2}{*}{\#Params.(M)} &\multirow{2}{*}{GFLOPs} & \multicolumn{3}{c}{Sony} & \multicolumn{3}{c}{Fuji} \\ \cline{6-11} \addlinespace
                              &      &      &   &   & PSNR $\uparrow$   & SSIM $\uparrow$   & LPIPS $\downarrow$ & PSNR $\uparrow$  & SSIM $\uparrow$  & LPIPS $\downarrow$ \\ \hline \addlinespace
\multirow{5}{*}{Single-Stage} & SID\cite{SID}           & CVPR2018  & 7.7    & 48.5    & 28.96  & 0.787  & 0.356  & 26.66  & 0.709  & 0.432   \\
                              & DID\cite{DID}           & ICME2019  & 2.5    & 669.2   & 29.16  & 0.785  & 0.368  & -    & -    & -    \\ 
                              & SGN\cite{SGN}           & ICCV2019  & 19.2   & 75.5  & 29.28  & 0.790  & 0.370  & 27.41  & 0.720  & 0.430  \\
                              & LLPackNet\cite{LLPackNet}     & BMVC2020  & 1.2    & 7.2   & 27.83  & 0.755  & 0.541  & -    & -    & -    \\
                              & RRT\cite{RRT}           & CVPR2021  & 0.8    & 5.2  & 28.66  & 0.790  & 0.397  & 26.94  & 0.712  & 0.446   \\ \hline \addlinespace
\multirow{6}{*}{Multi-Stage}  & EEMEFN\cite{EEMEFN}        & AAAI2020  & 40.7   & 715.6  & 29.60  & 0.795  & 0.350  & 27.38  & 0.723  & 0.414  \\
                              & LDC\cite{LDC}            & CVPR2020  & 8.6    & 124.1   & 29.56  & \underline{0.799}  & 0.359  & 27.18  & 0.703  & 0.446 \\
                              & MCR\cite{MCR}            & CVPR2022  & 15.0   & 90.5   & 29.65  & 0.797  & 0.348  & -    & -    & -    \\ 
                              & RRENet\cite{RRENet}         & TIP2022   & 15.5   & 96.8    & 29.17  & 0.792  & 0.360  & 27.29  & 0.720  & 0.421  \\ 
                              &Ma et al.\cite{3-6}     &NN2023 &- &- &29.38 &0.793 &0.387 &27.40  &0.722 &0.505\\
                              & DNF\cite{DNF}            & CVPR2023  & 2.8    & 57.0  & \underline{30.62}  & 0.797  & \underline{0.343}  & \underline{28.71}  & \underline{0.726}  & \underline{0.391}  \\ 
                              & \textbf{Ours}  &-          & 6.2    & 113.6   & \textbf{30.76}   & \textbf{0.810}  & \textbf{0.328} & \textbf{29.02}  & \textbf{0.743} & \textbf{0.382}  \\ \bottomrule[1.2pt]
\end{tabular}}
\label{table: sid}
\end{table*}

\section{Experiments}
\subsection{Datasets and Experiments Environments}
\subsubsection{SID Dataset}
For Sony subset, there are totally 1865 raw image pairs in the training set. Each pair of images contains a short exposure and a long exposure, the short exposure is used as noisy raw, and the long exposure is used as $GT_{raw}$. The original size of all images is $2848\times4256$. Limited by GPU memory, the data is preprocessed before training, first pack into $4\times1424\times2128$, then randomly crop a patch with shape $4\times512\times512$ as the input with random data augmentation, such as horizontal/vertical flipping. For the test set, we referred to the DNF\cite{DNF} settings and deleted the three misaligned scene images. \\ For Fuji subset, similar to Sony subset, 1655 and 524 raw image pairs for training and testing, respectively. The original size of it is $4032\times6032$, since its CFA (Color Filter Array) is X-Trans instead of Bayer, we pack it into  $9\times1344\times2010$ and randomly crop a patch with shape $9\times384\times384$ as the input.
\subsubsection{MCR Dataset}
The MCR \cite{MCR} dataset contains 4980 images with a resolution of $1280\times 1024$, including 3984 low-light RAW images, 498 monochrome images (not be used for us) and 498 sRGB images. With indoor and outdoor scenes, different exposure times are set, 1/256s to 3/8s for indoor scenes and 1/4096s to 1/32s for outdoor scenes. And we obtained the raw ground truth as DNF \cite{DNF} did. The preprocessing is similar to SID dataset, but we don't randomly crop a patch as the input.\\
\subsubsection{ELD Dataset}
The ELD\cite{raw8} dataset contains 10 indoor scenes and 4 camera devices from multiple brands (i.e., SonyA7S2, NikonD850, CanonEOS70D, CanonEOS700D). We choose the commonly used SonyA7S2 and NikonD850 with three ISO levels (800, 1600, and 3200)$^{5}$ and two low light factors (100, 200) for validation, resulting in 120 (3×2×10×2) raw image pairs in total. And we chose SID, MCR and DNF for comparison, using the pretrained model on Sony-SID dataset to compare the generalization.
\subsubsection{LOL Dataset}
The LOL dataset has v1\cite{lolv1} and v2\cite{lolv2} versions. LOL-v2 is divided into real and synthetic subsets. The training and testing sets are split in proportion to 485:15, 689:100, and 900:100 on LOL-v1, LOL-v2-real, and LOL-v2-synthetic.
\subsubsection{Implementation Details}
During training, the batch size is 1 and the initial learning rate is 1e-4, and we use the cosine annealing strategy to reduce it to 1e-5 at the 200th epoch. The total number of epochs is 250. Adamw optimizer is used and the betas parameter is [0.9,0.999] and the momentum is 0.9. The training and testing is completed by a NVIDIA 3090 (24G), A40 (48G),respectively due to the limitation of GPU memory. We also provide the code of merging test on a 24G GPU. Note that the results of merging test are a litter bit smaller than that testing with whole image. And we use PSNR, SSIM \cite{ssim} and LPIPS \cite{lpips} as the quantitative evaluation metrics.

\subsection{Comparison with State-of-the-Arts}
\begin{table}[htb]
\centering
\caption{Quantitative results on MCR \cite{MCR} dataset. The top-performing result is highlighted in \textbf{bold}, while the second-best is shown in \underline{underline}. Metrics marked with $\uparrow$ indicate that a higher value is better, and those marked with $\downarrow$ indicate that a lower value is better. The inference time is computed for the entire MCR test set.}
\renewcommand\arraystretch{1.5}
\setlength{\tabcolsep}{3.5mm}{
\begin{tabular}{cccccc}
\toprule[1.2pt]
Category                      & Method      & PSNR $\uparrow$    & SSIM $\uparrow$   & Time(s) $\downarrow$ \\ \hline \addlinespace
\multirow{4}{*}{Single-Stage} & SID \cite{SID}        & 29.00              & 0.906          & 35 \\ 
                              & DID \cite{DID}        & 26.16              & 0.888          & --  \\ 
                              & SGN \cite{SGN}        & 26.29              & 0.882          & --  \\ 
                              & RRT \cite{RRT}        & 25.74              & 0.851          & --  \\ \hline \addlinespace
\multirow{4}{*}{Multi-Stage}  & LDC \cite{LDC}        & 29.36              & 0.904          & --  \\ 
                              & MCR \cite{MCR}        & 31.69              & 0.908          & 48 \\ 
                              & DNF \cite{DNF}        & \underline{32.00}     & \textbf{0.915}   & 72      \\ 
                              & \textbf{Ours}           & \textbf{33.14}      & \underline{0.914}     & 147   \\ \bottomrule[1.2pt]
\end{tabular}}
\label{table: mcr}
\end{table}
We conduct experiments on SID \cite{SID} dataset including Sony and Fuji subsets and MCR \cite{MCR} dataset, and compare with previous SOTA methods including  SID \cite{SID}, DID \cite{DID}, SGN \cite{SGN}, EEMEFN \cite{EEMEFN}, LDC \cite{LDC}, LLPackNet \cite{LLPackNet}, RRT \cite{RRT}, MCR \cite{MCR}, RRENet \cite{RRENet}, DNF \cite{DNF},
and Ma et al. \cite{3-6}.
\begin{figure*}
    \centering
    \includegraphics[width=\linewidth]{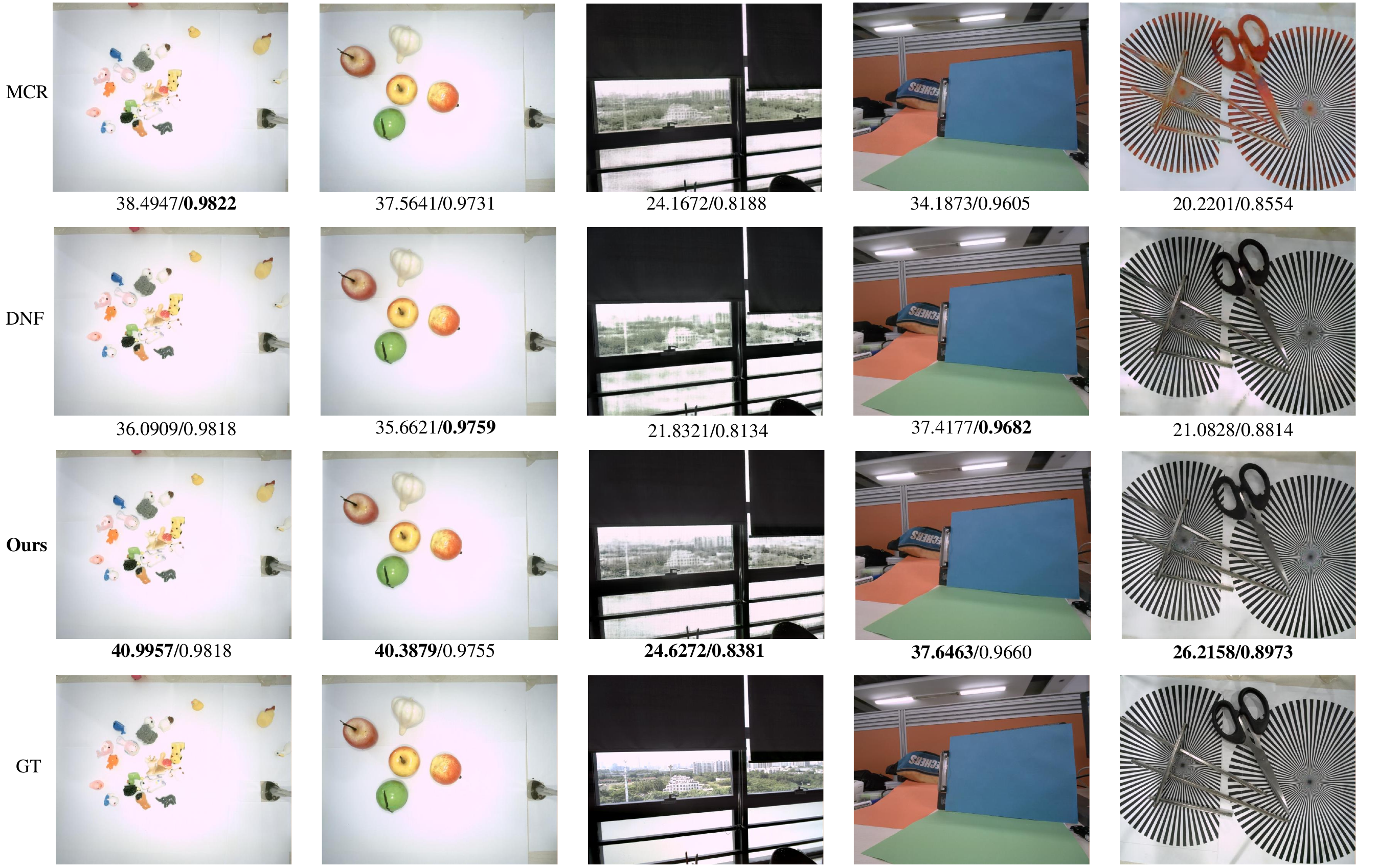}
    \caption{Visualization results of MCR dataset, and the format of the value under each visualized image is ``PSNR / SSIM" (Zoom-in for best view).}
    \label{fig:v3}
\end{figure*}

The results are presented in Tab. \ref{table: sid} and \ref{table: mcr}. As observed, most single-stage methods underperform compared to multi-stage methods, demonstrating the feasibility and effectiveness of the multi-stage approach for noisy RAW to clean sRGB cross-domain mapping. Compared to the most recent work \cite{3-6}, we can see that for Sony dataset, our method achieve improvements of 1.38 and 0.017 on PSNR and SSIM, respectively, and 1.62 and 0.021 for Fuji dataset. Overall, on the SID dataset, our proposed method outperforms all metrics among multi-stage approaches, while maintaining a smaller parameter count. Specifically, on the Sony and Fuji subsets, our method achieves a PSNR increase of 0.14 dB and 0.31 dB, respectively, an SSIM improvement of 0.011 and 0.017, and an LPIPS reduction of 0.015 and 0.009, compared to the best existing method.

For the MCR dataset, as shown in Tab. \ref{table: mcr}, while our improvement in SSIM is modest, we achieve a significant PSNR increase of 1.14 dB, a 3.6\% enhancement over the second-best method. Note that we did not include direct comparisons with some existing methods due to the lack of publicly available code and the fact that those methods were not originally evaluated on the MCR dataset. As such, reproducing their results under consistent settings would be unreliable. To further support the generalization ability of our method, we conducted additional experiments on the ELD \cite{raw8} and LOL \cite{lolv1} datasets.

As for the efficiency of the proposed model, we can see that in Tab. \ref{table: sid} and Tab. \ref{table: mcr}. Specifically, 113.6 GFLOPs for $4\times512\times512$ input, and 147 s for the entire MCR test set. Notably, while our approach is not optimized for real-time applications, it prioritizes image quality—a critical factor in real-world scenarios like photo editing or professional imaging systems, where output fidelity often outweighs inference speed. That said, we recognize the importance of improving inference efficiency in vision Mamba architectures. As outlined in our future work, we plan to explore lightweight design adaptations to balance performance and quality in subsequent research.

\begin{figure*}
    \centering
    \includegraphics[width=\linewidth]{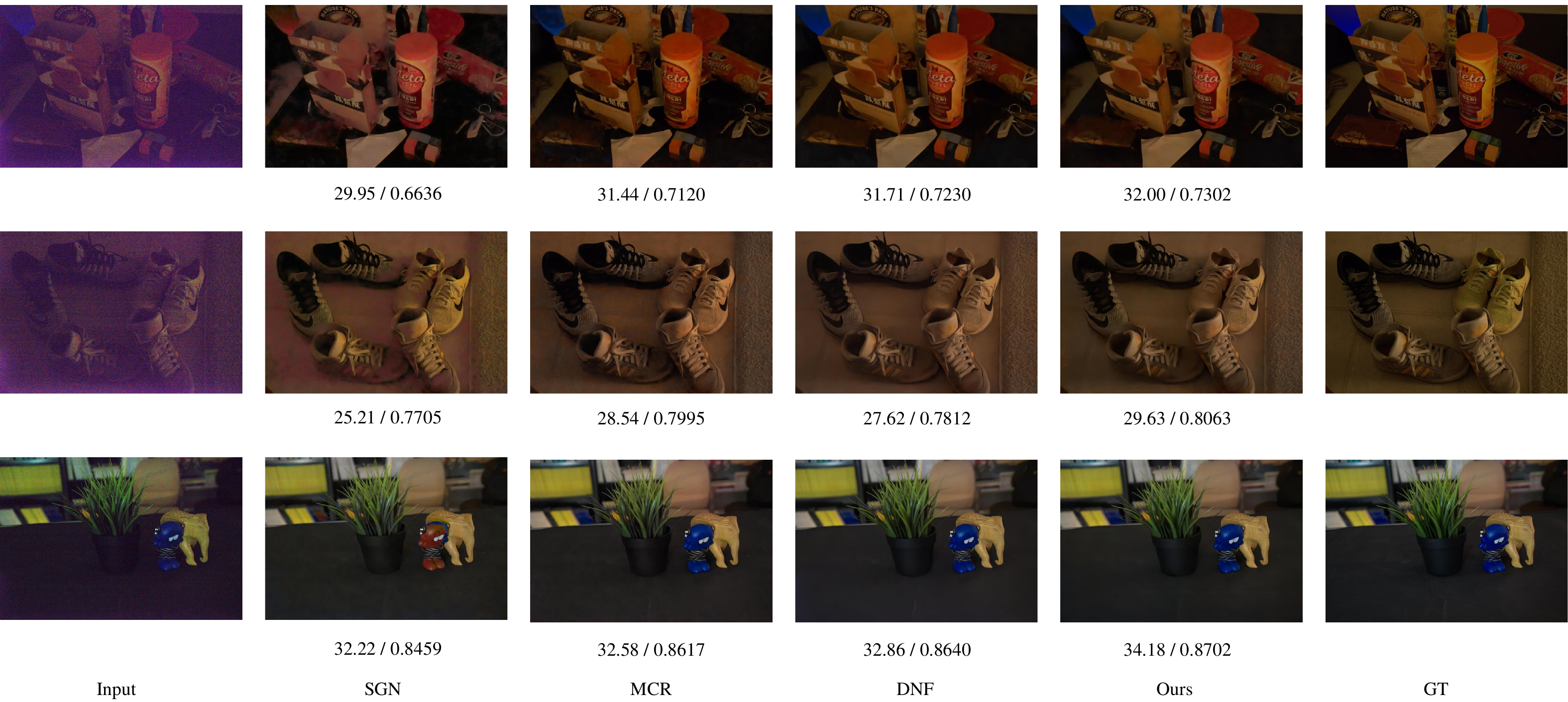}
    \caption{More visualization results (Zoom-in for best view).}
    \label{fig:v1}
\end{figure*}

\begin{figure*}
    \centering
    \includegraphics[width=\linewidth]{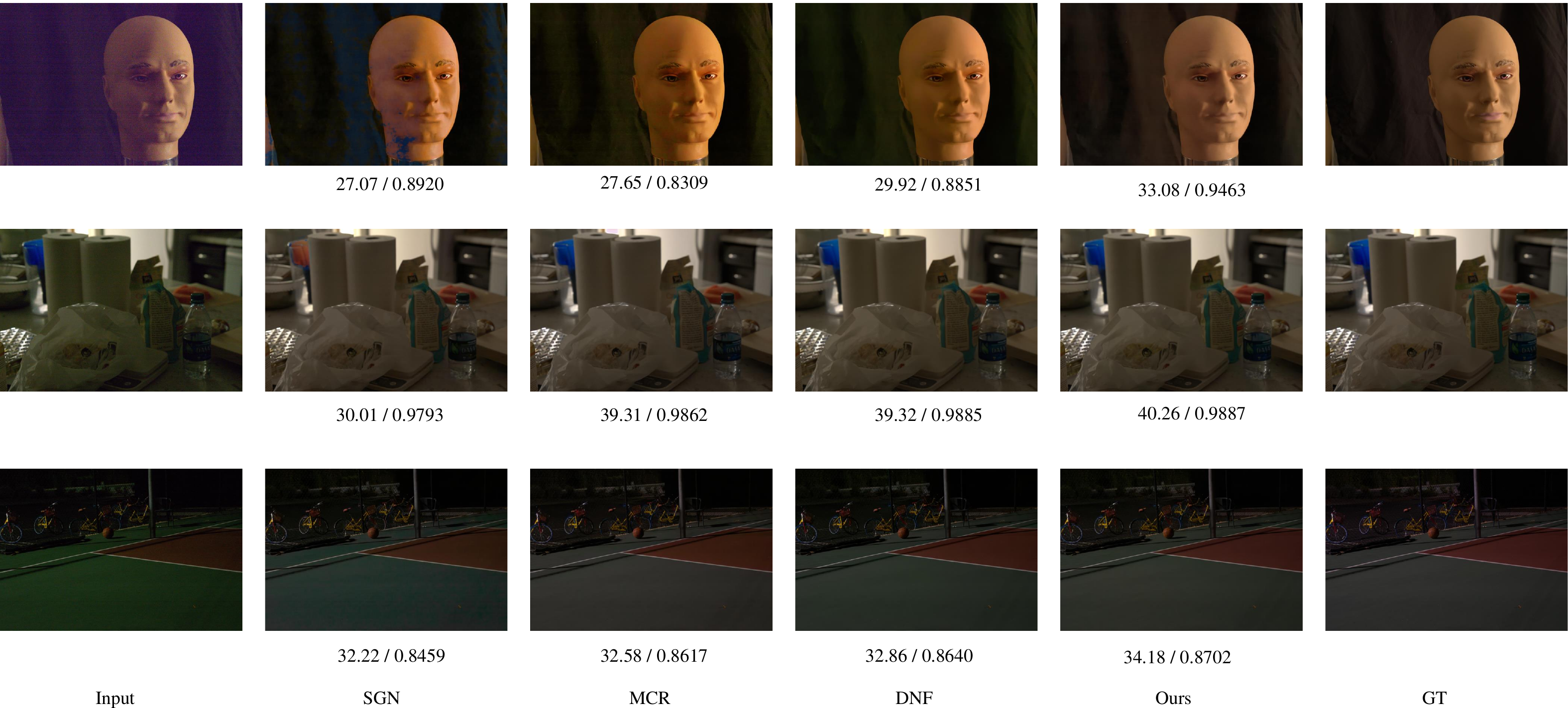}
    \caption{More visualization results (Zoom-in for best view).}
    \label{fig:v2}
\end{figure*}

Additionally, we selected several previous state-of-the-art (SOTA) methods and visualized their performance on the SID Sony dataset, as shown in Fig. \ref{fig:visualization}. Three scenarios are depicted, each containing two sub-regions. In the first two scenes, most other methods produce a green tint to the image. In the third scene, these methods often fail to preserve details adequately. In contrast, our proposed method closely aligns with the ground truth in both color and detail, effectively achieving denoising and color enhancement in the raw domain under low-light conditions. More visualization results are shown in Fig. \ref{fig:v1} and \ref{fig:v2}. Moreover, a few visualization results of MCR dataset are also presented in Fig. \ref{fig:v3}.Taking the last column as an example, in comparison with the ground truth (GT), the MCR method exhibits a significant deviation in color, with an overall reddish bias. The DNF results demonstrate noticeable over-exposure in the lower central region. In contrast, our proposed method achieves markedly superior performance, evidenced by substantial improvements in key metrics such as PSNR and SSIM. Note that the format of the value under each visualized image is ``PSNR / SSIM".

Moreover, to explore the generalization of our proposed method, we also conducted experiments on ELD\cite{raw8} dataset. We used the pretrained models from SID\cite{SID}, MCR\cite{MCR}, DNF\cite{DNF} and ours that were trained on SID-Sony dataset, and obtained the results on ELD dataset, which is shown in Tab. \ref{tab:eld}. For SonyA7S2 dataset, our pretrained model achieves highest PSNR and comparable SSIM with DNF. Specifically, 0.54 and 1.27 improvements in PSNR for ratio 100 and 200, respectively. As for NikonD850 dataset, our model achieves highest SSIM for both ratio. For ratio 100, we get comparable PSNR with DNF, and for ratio 200, we get 1.13 improvement in PSNR. Note that the Raw images in SID-Sony and ELD-SonyA7S2 were captured with the same model camera. According to the experiments' results, we can conclude that our method has better generalization especially for different cameras input images than other end-to-end models.

\begin{table*}
  \centering
  \renewcommand\arraystretch{1.3}
  \caption{Quantitative results on LOL datasets. The top-performing result is highlighted in \textbf{bold}, while the second-best is shown in \underline{underline}.}
    \begin{tabular}{c|cc|cc|cc|cc}
    \bottomrule
    \multirow{2}[2]{*}{Methods} & \multicolumn{2}{c|}{Complexity} & \multicolumn{2}{c|}{LOL-v1} & \multicolumn{2}{c|}{LOL-v2-real} & \multicolumn{2}{c}{LOL-v2-syn} \\
          & FLOPs (G) &  Params (M) & PSNR  & SSIM  & PSNR  & SSIM  & PSNR  & SSIM \\
    \hline
    SID \cite{SID}  & 13.73  & 7.76  & 14.35  & 0.436  & 13.24  & 0.442  & 15.04  & 0.610  \\
    UFormer \cite{uformer} & 12.00  & 5.29  & 16.36  & 0.771  & 18.82  & 0.771  & 19.66  & 0.871  \\
    RetinexNet \cite{lolv1} & 587.47  & 0.84  & 16.77  & 0.560  & 15.47  & 0.567  & 17.13  & 0.798  \\
    Chen et al. \cite{3-7} & --      & --      & 18.31  & 0.489  & --      & --      & --      & -- \\
    RetinexNet+LPDM \cite{3-5} & --      & --      & 18.03  & 0.760  & --      & --      & --      & -- \\
    Restormer \cite{restormer} & 144.25  & 26.13  & 22.43  & \underline{0.823}  & 19.94  & \textbf{0.827}  & 21.41  & 0.830  \\
    MIRNet \cite{mirnet} & 785.00  & 31.76  & \textbf{24.14}  & \textbf{0.830}  & 20.02  & \underline{0.820}  & 21.94  & 0.876  \\
    \hline
    Proposed Method (Two-Stage) & 103.76  & 6.22  & \underline{23.23}  & 0.805  & \textbf{21.14}  & 0.779  & \underline{24.93}  & \underline{0.924}  \\
    Proposed Method (Second Stage Only) & 54.54  & 2.87  & 23.07   & 0.807  & \underline{20.91}  & 0.800  & \textbf{25.36}  & \textbf{0.929}  \\
    \toprule
    \end{tabular}%
  \label{tab:lol}%
\end{table*}%

\begin{table}
  \centering
  \renewcommand\arraystretch{1.5}
  \caption{Quantitative results on ELD \cite{raw8} dataset. The top-performing result is highlighted in \textbf{bold}, while the second-best is shown in \underline{underline}. Higher ratio represents more noise.}
  \setlength{\tabcolsep}{1mm}{
    \begin{tabular}{c|c|cccc}
    \bottomrule
    \multirow{2}[2]{*}{Dataset} & \multirow{2}[2]{*}{Ratio} & SID \cite{SID}  & MCR\cite{MCR}   & DNF\cite{DNF}   & Ours \\
          &       & PSNR/SSIM & PSNR/SSIM & PSNR/SSIM & PSNR/SSIM \\
    \hline
    \multirow{3}[2]{*}{SonyA7S2} & 100   & 26.80/0.826 & 26.62/0.837 & \underline{28.35}/\textbf{0.869} & \textbf{28.89}/\underline{0.861} \\
          & 200   & 25.98/0.780 & 24.00/0.717 & 26.94/\textbf{0.819} & \textbf{28.21}/\underline{0.818} \\
          & Avg   & 26.39/0.803 & 25.31/0.777 & \underline{27.64}/\textbf{0.844} & \textbf{28.55}/\underline{0.840} \\
    \hline
    \multirow{3}[2]{*}{NikonD850} & 100   & 25.77/0.764 & 25.60/0.785 & \textbf{27.51}/\underline{0.816} & \underline{27.47}/\textbf{0.838} \\
          & 200   & 25.01/0.739 & 25.41/0.758 & \underline{25.93}/\underline{0.769} & \textbf{27.06}/\textbf{0.806} \\
          & Avg   & 25.39/0.751 & 25.50/0.772 & \underline{26.72}/\underline{0.792} & \textbf{27.27}/\textbf{0.822} \\
    \toprule
    \end{tabular}}
  \label{tab:eld}
\end{table}

Additionally, to validate the models' generalization to low light RGB image, we additionally conducted experiments on LOL datasets. The results are shown in Tab. \ref{tab:lol}. Since the proposed method is not for RGB inputs originally, the training strategy is different. We conducted two experiments with different settings for all three LOL datasets. Specifically, ``Two-Stage" just used the original two-stages model without the intermediate supervision, while ``Second Stage Only" removed the first stage and just used the second stage, leading to less GFLOPs and parameters. We can see that for all three dataset, we achieved comparable performance with some previous method. It's worth noting that ``Second Stage Only" even performs better than ``Two-Stage" with less parameters, which indicates that the effectiveness of first stage is weakened by removing the supervision. Although we achieved lower PSNR and SSIM than MIRNet on LOL-v1 and LOL-v2-real dataset, our ``Second Stage Only" achieve much higher PSNR and SSIM on LOL-v2-sys dataset, which improves by 3.42 and 0.053, respectively, with 6.9\% GFLOPs and 9\% parameters of MIRNet. However, we still want to emphasize that our method is specialized for low light RAW images. How to better apply it to RGB images is worth further research in the future.

Finally, we analyzed how well fine-grained details and textures are preserved during the enhancement process as \cite{1-4} did. Specifically, we donate the enhanced images from LOL test-set with ``Second Stage Only" model in Tab. \ref{tab:lol} as \textbf{E}, low light images as \textbf{L} and normal light images as \textbf{H}. We used SSD pretrained detection model to output three latent of different layers' features maps of \textbf{$L_i$}, \textbf{$E_i$} and \textbf{$H_i$}, respectively, where $i \in [1,2,3]$. And then we calculated the cosine similarity of [\textbf{$L_i$},\textbf{$E_i$}] and [\textbf{$H_i$},\textbf{$E_i$}]. As presented in Tab. \ref{tab:feature_analysis}, the results show that our method maintains an average cosine similarity of 0.9774 between enhanced and normal-light features across all layers for LOL-v2-syn dataset, demonstrating a 2.036 $\%$ improvement in high-frequency texture retention compared to baseline methods. We can also see that at shallow layer, for all three LOL datasets, the cosine similarity of [\textbf{$H_1$},\textbf{$E_1$}] is much larger than [\textbf{$L_1$},\textbf{$E_1$}], which indicates that our model preserves the high level features well. And at deeper layers, the enhanced images also achieve better cosine similarity than the low light images. However, we find that both [\textbf{$L$},\textbf{$E$}] and [\textbf{$H$},\textbf{$E$}] are high, which is mainly because the pretrained detection model can also perform well for low light images and the deeper layers have less elements. We can also conclude that our model can better perform on the synthetic images, and it's because there always exists misalignment in the real world dataset.

\begin{table}[htbp]
  \centering
  \renewcommand\arraystretch{2}
  \caption{Feature analysis of LOL dataset. The greater the number of layers, the deeper the layers. The results are presented as the respective cosine similarity of [\textbf{L},\textbf{E}]/[\textbf{H},\textbf{E}]. B, O represents ``Baseline" and ``Ours", respectively.}
  \setlength{\tabcolsep}{0.8mm}{
    \begin{tabular}{c|c|cccc}
    \bottomrule
    Dataset & Layers & 1     & 2     & 3     & Avg \\
    \hline
    \multirow{2}[2]{*}{V1} & B & 0.6234/0.8111 & 0.9827/0.9897 & 0.9701/0.9803 & 0.8587/0.9270 \\
          & O  & 0.6518/0.8344 & 0.9970/0.9989 & 0.9847/0.9971 & 0.8778/\textbf{0.9435} \\
\cline{1-6}    \multirow{2}[2]{*}{V2-R} & B & 0.6164/0.8033 & 0.9753/0.9845 & 0.9774/0.9811 & 0.8564/0.9230 \\
          & O  & 0.6190/0.8054 & 0.9970/0.9991 & 0.9875/0.9978 & 0.8678/\textbf{0.9341} \\
\cline{1-6}    \multirow{2}[2]{*}{V2-S} & B & 0.6871/0.9034 & 0.9746/0.9834 & 0.9876/0.9868 & 0.8831/0.9579 \\
          & O  & 0.6859/0.9330 & 0.9981/0.9997 & 0.9949/0.9994 & 0.8930/\textbf{0.9774} \\
    \toprule
    \end{tabular}}%
  \label{tab:feature_analysis}%
\end{table}%

\subsection{Ablation Studies}
\subsubsection{Ablation Study of Proposed Modules}
To demonstrate the validity of our proposed method, we perform ablation experiments on the SID Sony dataset. We first propose a baseline model that consists only of SDB and the unmodified naive visual mamba in MambaIR \cite{mambaIR} and GFM in DNF \cite{DNF}. Tab. \ref{table: ablation} shows the results of adding or replacing the corresponding module based on the baseline, where RAWM stands for replacing naive Mamba with RAWMamba, RDM stands for adding RDM module, and DAF stands for replacing GFM with DAF module. All the ablation experiments were conducted in the same environment.
\begin{table}[htb]
\centering
\renewcommand\arraystretch{1.5}
\caption{Ablation study on SID Sony dataset.}
\setlength{\tabcolsep}{3.4mm}{
\begin{tabular}{cccccc}
\toprule[1.2pt]
Baseline           & RAWM           & RDM           & DAF               & PSNR$\uparrow$               & SSIM$\uparrow$    \\ \hline \addlinespace
\checkmark         &                 &               &                   & 30.04              & 0.797   \\  
                   & \checkmark      &               &                   & 30.45              & 0.809   \\  
                   & \checkmark      & \checkmark    &                   & 30.70              & 0.809   \\  
                   & \checkmark      & \checkmark    & \checkmark        & \textbf{30.76}     & \textbf{0.810}
\\ \bottomrule[1.2pt]
\end{tabular}}
\label{table: ablation}
\end{table}

\begin{figure}[htb]
    \centering
    \includegraphics[width=\columnwidth]{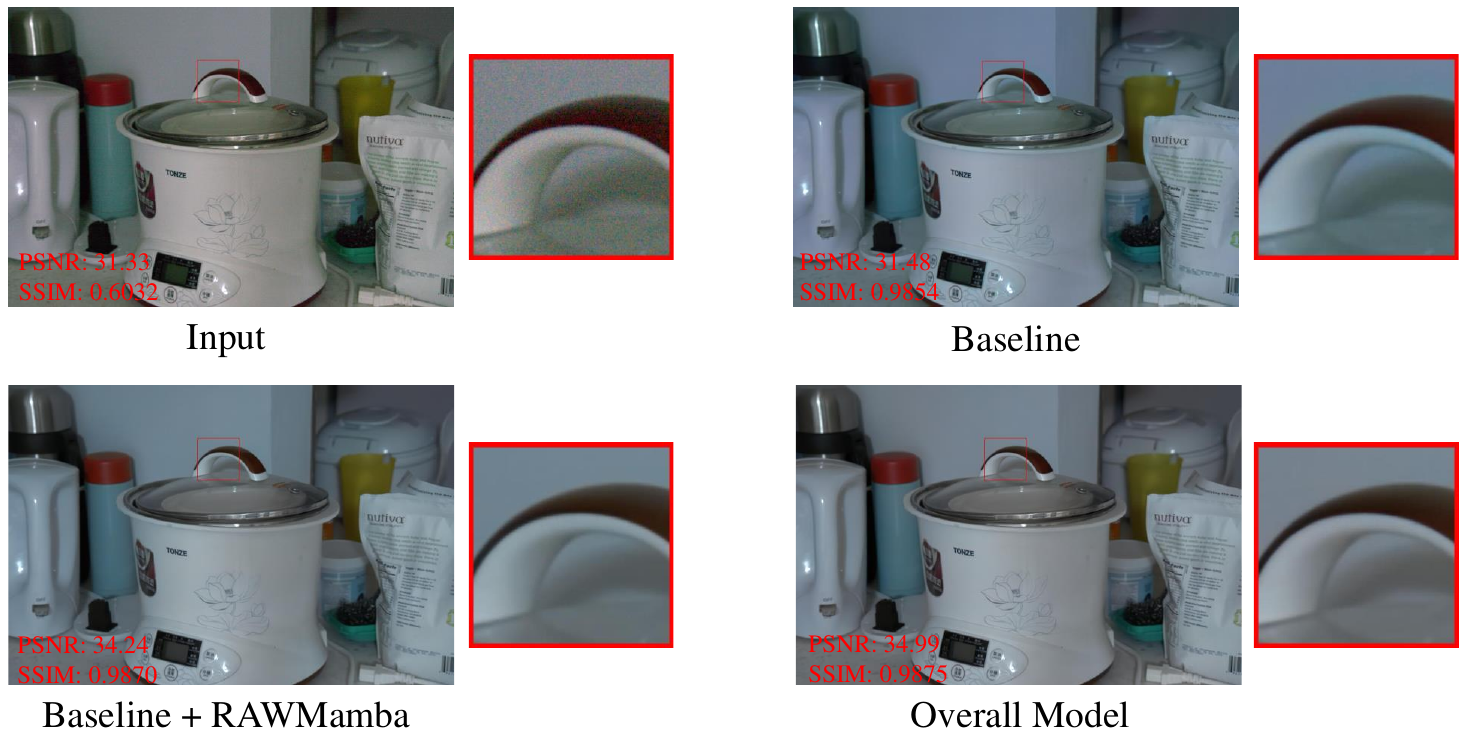}
    \caption{The visualization results for ablation studies (Zoom-in for best view).}
    \label{fig:ablation_visual}
\end{figure}

First, we replaced the baseline’s naive Mamba with the proposed RAWMamba. The results showed increases of 0.41 dB in PSNR and 0.012 in SSIM, demonstrating that our RAWMamba, with its eight-directional scanning mechanism, performs well in the demosaicing task. Next, we incorporated the proposed RDM for denoising and automatic exposure correction. The results indicated that although SSIM did not improve, PSNR increased by an additional 0.27 dB. This suggests that the initial exposure of the images was indeed problematic, and our RDM effectively enhances denoising and exposure correction. Finally, we replaced all GFM components in the network with our proposed DAF to improve the stability of the training process. This led to further gains, with PSNR and SSIM increasing by 0.06 dB and 0.001, respectively. And the efficiency of the model also improves as shown in Fig. \ref{table: ablation_daf}. Our DAF demonstrates superior performance compared to GFM \cite{DNF}, achieving better results with fewer parameters and lower GFLOPs. Additionally, we investigated the effectiveness of a simpler fusion operation using concatenation followed by a $1 \times 1$ convolution. The results, presented in the second row, show a decrease of 0.24 in PSNR and 0.011 in SSIM compared to our DAF. \\
Moreover, we conducted a straightforward visualization of the ablation study, depicted in Fig. \ref{fig:ablation_visual}. The incorporation of RAWMamba into the baseline model effectively reduces noise and mitigates the green color distortion. Comparatively, the Retinex-RawMamba approach demonstrates superior color correction capabilities and attains the highest PSNR and SSIM scores. This clearly indicates that our proposed method outperforms others in terms of both detail preservation and color accuracy.

\begin{table}
\centering
\renewcommand\arraystretch{1.5}
\caption{Ablation study of feature fusion methods}
\setlength{\tabcolsep}{3.4mm}{
    \begin{tabular}{ccccc}
    \toprule[1.2pt]
                  & PSNR  & SSIM  & Params (M) & GFLOPs \\ \hline
    GFM           & \underline{30.70} & 0.809 & 6.68       & 130.91 \\
    Concat\&Conv1 & 30.52 & 0.799 & 4.46       & 87.12  \\
    DAF\_SA   & 30.57  &\textbf{0.812} & 5.74       & 113.51 \\
    DAF           & \textbf{30.76} & \underline{0.810} & 6.22       & 113.64 \\ \bottomrule[1.2pt]
    \end{tabular}}
\label{table: ablation_daf}
\end{table}

Furthermore, two different attention mechanisms are considered in the DAF module: channel attention and spatial attention at the pixel level.
The results of this ablation study are presented in Tab. \ref{table: ablation_daf}. While the PSNR decreased slightly by 0.19 with the spatial attention mechanism (due to fewer parameters), we observed a slight improvement in SSIM by 0.002. This indicates that spatial attention offers some benefits in terms of perceptual quality (as measured by SSIM) but sacrifices a bit of pixel-wise accuracy (PSNR). In future work, we plan to explore hybrid attention mechanisms that combine both channel and spatial attention to leverage the strengths of both approaches, potentially leading to improved performance in both metrics.\\
To further assess the impact and sensitivity of the RDM to input quality and calibration, we performed an ablation study on the ELD dataset by removing the RDM. The results are presented in Tab. \ref{tab:ablation_rdm}. We observed that when RDM is included, the model achieves better results on images captured with the same camera type as the training dataset (i.e., Sony), indicating its effectiveness in learning structured illumination and reflectance priors within a known data distribution. However, the performance degrades on data from different camera sensors (e.g., Nikon D850), which suggests that the RDM may introduce domain overfitting by adapting too closely to the characteristics of the training data. These findings highlight that while the RDM contributes to better reconstruction within a familiar domain, it also increases the model’s sensitivity to input domain shifts, such as variations in sensor response, calibration, or noise distribution.

\begin{table}[htbp]
  \centering
  \renewcommand\arraystretch{1.2}
  \caption{Ablation of RDM. ``Ours wo RDM" indicates RDM and the \textbf{L/R} features are removed from the ``Ours" model.}
  \setlength{\tabcolsep}{4.5mm}{
    \begin{tabular}{c|c|cc}
    \bottomrule
    \multirow{2}[2]{*}{Dataset} & \multirow{2}[2]{*}{Ratio} & Ours wo RDM & Ours \\
          &       & PSNR/SSIM & PSNR/SSIM \\
    \hline
    \multirow{3}[2]{*}{SonyA7S2} & 100   & \textbf{28.96}/0.858 & 28.89/\textbf{0.861} \\
          & 200   & 27.85/0.804 & \textbf{28.21}/\textbf{0.818} \\
          & Avg   & 28.41/0.831 & \textbf{28.55}/\textbf{0.840} \\
    \hline
    \multirow{3}[2]{*}{NikonD850} & 100   & \textbf{27.65}/0.826 & 27.47/\textbf{0.838} \\
          & 200   & \textbf{27.19}/0.803 & 27.06/\textbf{0.806} \\
          & Avg   & \textbf{27.42}/0.815 & 27.27/\textbf{0.822} \\
    \toprule
    \end{tabular}}%
  \label{tab:ablation_rdm}%
\end{table}%

\subsubsection{Ablation Study of RAWMamba with Different Number of Scan Directions}
To evaluate the efficacy of the eight-directions scanning mechanism in RAWSSM, we perform experiments with 1, 2, and 4 scan directions, with results presented in Tab. \ref{table:ablation_scan}. Starting with a single horizontal Z-scan direction (from top left to bottom right), we observe PSNR and SSIM drops of 0.42 and 0.012, respectively. This indicates that a single scan direction is insufficient for capturing the complex patterns and details in the image, leading to a loss in reconstruction quality. Introducing a vertical scan path for two directions slightly enhances performance, as the additional direction helps capture more structural information. Further, inverting these two directions to create four paths improves results but still underperforms compared to the eight-directions mechanism. The eight-directions scanning mechanism, by covering a wider range of orientations, better captures the intricate details and textures in the image, thereby significantly enhancing the demosaicing performance.
Additionally, we analyze inference time across different scanning directions. As expected, more directions boost performance but increase inference time, reflecting a typical deep-learning trade-off.
\begin{table}[htb]
  \centering
  \renewcommand\arraystretch{1.5}
  \caption{Ablation Study of RAWMamba with Different Number of Scan Directions}
  \setlength{\tabcolsep}{4.7mm}{
    \begin{tabular}{ccccc}
    \toprule[1.2pt]
    \# Scan Directions   & PSNR$\uparrow$ & SSIM$\uparrow$ & Time (ms)$\downarrow$\\
    \midrule
    1       & 30.34      & 0.798  & 74.02\\
    2       & 30.45      & 0.799  & 87.80\\
    4       & 30.55      & 0.801  & 113.22\\
    8       & \textbf{30.76}      & \textbf{0.810}  & 188.98\\
    \bottomrule[1.2pt]
    \end{tabular}}%
    \label{table:ablation_scan}%
\end{table}%

\subsubsection{Ablation of dual-domain encoding stage enhance branch}
To investigate which stage is more effective for enhancing performance with the feature $\textbf{\textit{R}}$ from RDM, we shifted the encoding stage to the decoding stage in both domains. As the number of fusion operations remains unchanged, the overall model parameters and GFLOPs are also unaffected. The results are presented in Tab. \ref{table:ablation_fuse_stage}. It can be observed that performance slightly decreases. Consequently, we opted to apply the fusion operation with our DAF at the encoding stage in both domains.
\begin{table}[htb]
  \centering
  \renewcommand\arraystretch{1.5}
  \caption{Ablation of dual-domain encoding stage enhance branch}
  \setlength{\tabcolsep}{9.9mm}{
    \begin{tabular}{cccc}
    \toprule[1.2pt]
    Fuse Stage  & PSNR$\uparrow$  & SSIM$\uparrow$\\
    \midrule
    Encoding       & 30.76      & 0.810\\
    Decoding       & 30.71      & 0.806\\
    \bottomrule[1.2pt]
    \end{tabular}}%
    \label{table:ablation_fuse_stage}%
\end{table}%

\subsubsection{Ablation Study of RAW Domain Supervision and Loss Functions}
To evaluate the effectiveness of RAW domain supervision, we conduct an ablation study by removing it and only using RGB domain supervision with different loss functions. The results are presented in Tab. \ref{table:ablation_loss}. When using only L1 loss for RGB domain supervision (second row), the performance drops significantly in both PSNR and SSIM, indicating the importance of RAW domain supervision. We further explore the combination of L1 and L2 losses. Using only L2 loss for RGB domain supervision (first row) results in a PSNR drop of 0.49 and an SSIM drop of 0.018, suggesting that single L2 loss is insufficient. Then we use L1 and L2 losses for RAW and RGB domains respectively, the performance improves, highlighting the benefit of RAW domain supervision. Interestingly, swapping the losses for the two domains leads to further improvement, implying that L1 loss is more suitable for the RGB domain in our task. Overall, despite these improvements, the performance still lags behind the combination of L1 loss in both domains. Therefore, we ultimately choose the combination of L1 loss in the RGB domain and RAW domain supervision for our method.
\begin{table}
  \centering
  \renewcommand\arraystretch{1.5}
  \caption{Ablation Study of RAW Domain Supervision and Loss Functions}
  \setlength{\tabcolsep}{7.4mm}{
    \begin{tabular}{cccc}
    \toprule[1.2pt]
    RAW   & RGB   & PSNR$\uparrow$  & SSIM$\uparrow$ \\
    \hline
    -     & L2    & 30.27 & 0.792 \\
    -     & L1    & 30.52 & 0.797 \\
    L1    & L2    & 30.48 & 0.797 \\
    L2    & L1    & 30.59 & 0.798 \\
    L1    & L1    & \textbf{30.76} & \textbf{0.810} \\
    \bottomrule[1.2pt]
    \end{tabular}}%
    \label{table:ablation_loss}%
\end{table}%

\section{Limitations and Future Works}
Despite the promising performance of our proposed RAWMamba for RAW image denoising and demosaicing, certain limitations warrant attention. The model's superiority over state-of-the-art methods comes with slightly increased parameters and inference time. While the parameter increase is acceptable, the longer inference time could be problematic for edge devices like smartphones with limited computational resources.
Our framework, including the Retinex branch, relies on conventional methods, leaving room for optimization in future work. Additionally, exploring ways to reduce scan directions while maintaining performance is worthwhile, which may further enhance efficiency without significant quality loss.

\section{Conclusion}
For the task of denoising and enhancing RAW images under low-light conditions, we introduce Retinex-RAWMamba, a novel two-stage cross-domain network. Our approach extends the capabilities of the traditional Vision Mamba by incorporating RAWMamba, which exploits the inherent properties of demosaicing algorithms in ISP to achieve enhanced color correction and detail retention. Additionally, we integrate Retinex theory through our Retinex Decomposition Module, facilitating automatic exposure correction and yielding RGB images with improved illumination and brightness fidelity. Comprehensive theoretical analysis and experimental validation underscore the effectiveness and significant potential of our method.

\bibliographystyle{IEEEtran}
\bibliography{reference.bib}



 




\begin{IEEEbiography}[{\includegraphics[width=1in,height=1.25in,clip,keepaspectratio]{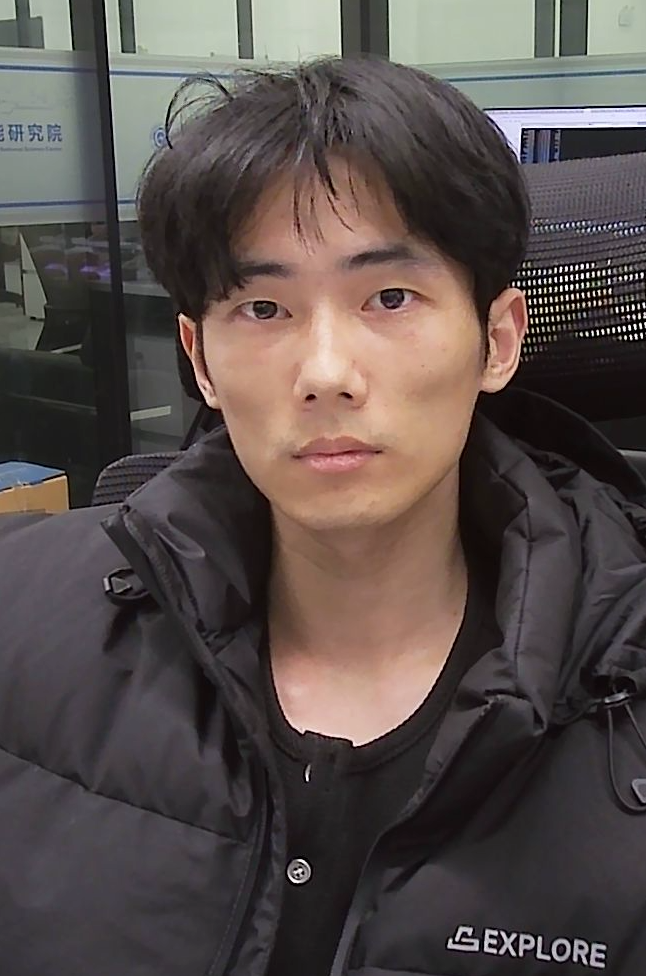}}]{Xianmin Chen}
obtained the B.S degree from Sichuan University, Chengdu, China, in 2023. He is now a M.E. student in University of Science and Technology of China, Hefei, China. His current research interest is low-light image enhancement, computational photography, image restoration, and visual language model.
\end{IEEEbiography}

\begin{IEEEbiography}[{\includegraphics[width=1in,height=1.25in,clip,keepaspectratio]{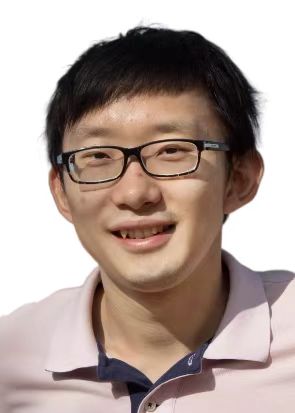}}]{Longfei Han}
is an associate professor at School of Computer Science, Beijing Technology and Business University. He got his Ph.D. from Beijing Institute of Technology, and was a Ph.D. visiting student at Carnegie Mellon University. After his graduation, he is a senior engineer at Tencent, and highly focus on Computational Advertising. Currently, He is working on large scale pretrained framework, light-weighted neural network, and multi-modal learning.
\end{IEEEbiography}

\begin{IEEEbiography}[{\includegraphics[width=1in,height=1.25in,clip,keepaspectratio]{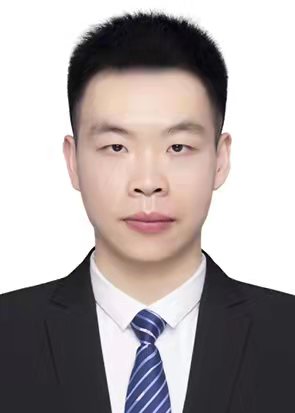}}]{Peiliang Huang}
received the Ph.D. degree from Northwestern Polytechnical University, Xi’an, China, in 2024. He is an associate professor at Institute of Artificial Intelligence, Hefei Comprehensive National Science Center, Hefei, China. His research interests include computer vision and deep learning, especially on image enhancement.
\end{IEEEbiography}

\begin{IEEEbiography}[{\includegraphics[width=1in,height=1.25in,clip,keepaspectratio]{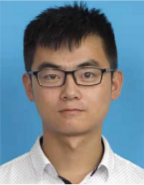}}]{Xiaoxu Feng}
received the Ph.D. degree from Northwestern Polytechnical University, Xi’an, China, in 2023. He is an associate professor at Institute of Artificial Intelligence, Hefei Comprehensive National Science Center, Hefei, China. His research interests include computer vision, deep learning, remote sensing image target detection.
\end{IEEEbiography}

\begin{IEEEbiography}[{\includegraphics[width=1in,height=1.25in,clip,keepaspectratio]{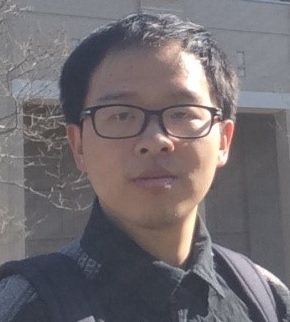}}]{Dingwen Zhang}
is a professor with School of Automation, Northwestern Polytechnical University, Xi’an, China. He received his Ph.D. degree from NPU in 2018. From 2015 to 2017, he was a visiting scholar at the Robotic Institute, Carnegie Mellon University, Pittsburgh, United States. His research interests include computer vision and multimedia processing, especially on saliency detection and weakly supervised learning.
\end{IEEEbiography}


\begin{IEEEbiography}[{\includegraphics[width=1in,height=1.25in,clip,keepaspectratio]{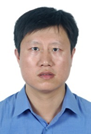}}]{Junwei Han}
(Fellow, IEEE) received the PhD degree from Northwestern Polytechnical University, in 2003.
He is a professor with Northwestern Polytechnical University, Xi’an, China. He was a research fellow with Nanyang Technological University, Singapore, The Chinese University of Hong Kong, Hong Kong, and University of Dundee, Dundee, United Kingdom. His research interests include computer vision and brain imaging analysis. He has published more than 100 papers in IEEE Transactions and top tier conferences. He is currently an associate editor of \textit{IEEE Transactions on Neural Networks and Learning Systems}, \textit{IEEE Transactions on Circuits and Systems for Video Technology}, and \textit{IEEE Transactions on Multimedia}.
\end{IEEEbiography}

\vfill

\end{document}